\documentclass[10pt,twocolumn,letterpaper]{article}

\usepackage[cvprfinal]{cvpr}      
\usepackage{url}
\usepackage{booktabs}
\usepackage{multicol}
\usepackage{multirow}
\usepackage{makecell}
\usepackage{diagbox}
\usepackage{wrapfig}
\usepackage{tabularx}
\usepackage{array}
\usepackage{graphicx}
\usepackage{marvosym}
\usepackage{xcolor}
\definecolor{cvprblue}{rgb}{0.21,0.49,0.74}
\usepackage[pagebackref,breaklinks,colorlinks,allcolors=cvprblue]{hyperref}

\def\paperID{14751} 
\def\confName{CVPR}
\def\confYear{2026}

\title{Video2LoRA: Unified Semantic-Controlled Video Generation via Per-Reference-Video LoRA}

\author{
Zexi Wu\textsuperscript{1},
\quad Baolu Li\textsuperscript{1},
\quad
Jing Dai\textsuperscript{1}, 
 \quad
Yiming Zhang\textsuperscript{1}, \quad
 \\
Yue Ma\textsuperscript{2}$^\text{\Letter}$, 
\quad
Qinghe Wang\textsuperscript{1}$^\text{\Letter}$,
\quad 
Xu Jia\textsuperscript{1}, \quad 
Hongming Xu\textsuperscript{1}$^\text{\Letter}$\\
\normalsize$^{1}$Dalian University of Technology ~~
\normalsize$^{2}$HKUST~~
\\[4pt]
\normalsize \href{https://github.com/BerserkerVV/Video2LoRA/}{\textcolor{blue}{\textit{https://github.com/BerserkerVV/Video2LoRA}}}
}

\begin{document}
\twocolumn[{%
\renewcommand\twocolumn[1][]{#1}%
\maketitle
\includegraphics[width=1\linewidth]{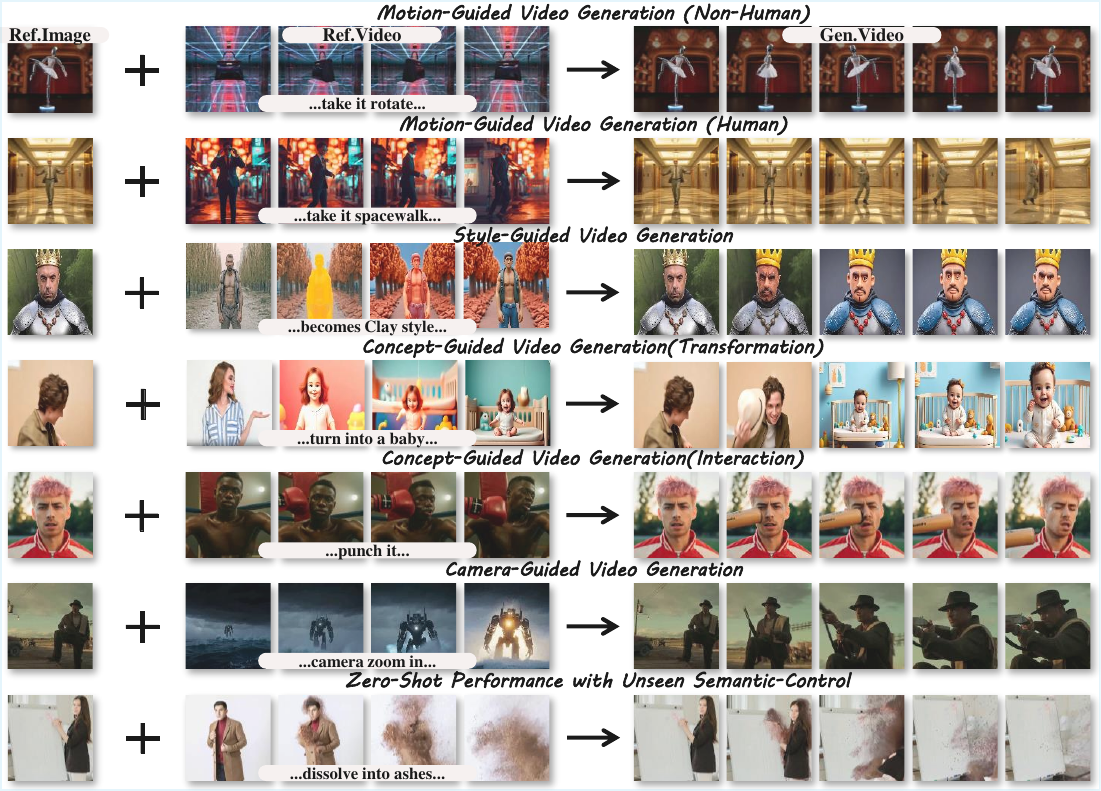}

\captionof{figure}{Video2LoRA is a unified framework for semantic-controllable video generation. It takes a reference video containing the desired semantics as input and employs a HyperNetwork to generate lightweight, semantic-specific LoRA modules. By integrating these adaptive components into a frozen video diffusion backbone, Video2LoRA achieves high-quality video generation in both within-domain and out-of-domain scenarios. \vspace{1em}}
\label{fig:teaser}
}]
\let\thefootnote\relax\footnotetext{
    \hspace{5pt}$^\text{\Letter}$ Corresponding Author\hspace{3pt}
} 
\begin{abstract}
Achieving semantic alignment across diverse video generation conditions remains a significant challenge. Methods that rely on explicit structural guidance often enforce rigid spatial constraints that limit semantic flexibility, whereas models tailored for individual control types lack interoperability and adaptability. These design bottlenecks hinder progress toward flexible and efficient semantic video generation. To address this, we propose \textbf{Video2LoRA}, a scalable and generalizable framework for semantic-controlled video generation that conditions on a reference video. Video2LoRA employs a lightweight hypernetwork to predict personalized LoRA weights for each semantic input, which are combined with auxiliary matrices to form adaptive LoRA modules integrated into a frozen diffusion backbone. This design enables the model to generate videos consistent with the reference semantics while preserving key style and content variations, eliminating the need for any per-condition training. Notably, the final model weights less than 150MB, making it highly efficient for storage and deployment. Video2LoRA achieves coherent, semantically aligned generation across diverse conditions and exhibits strong zero-shot generalization to unseen semantics.



\end{abstract}    
\begin{figure*}[t]
\centering
\includegraphics[width=1\linewidth, height=0.6\linewidth, keepaspectratio=false]{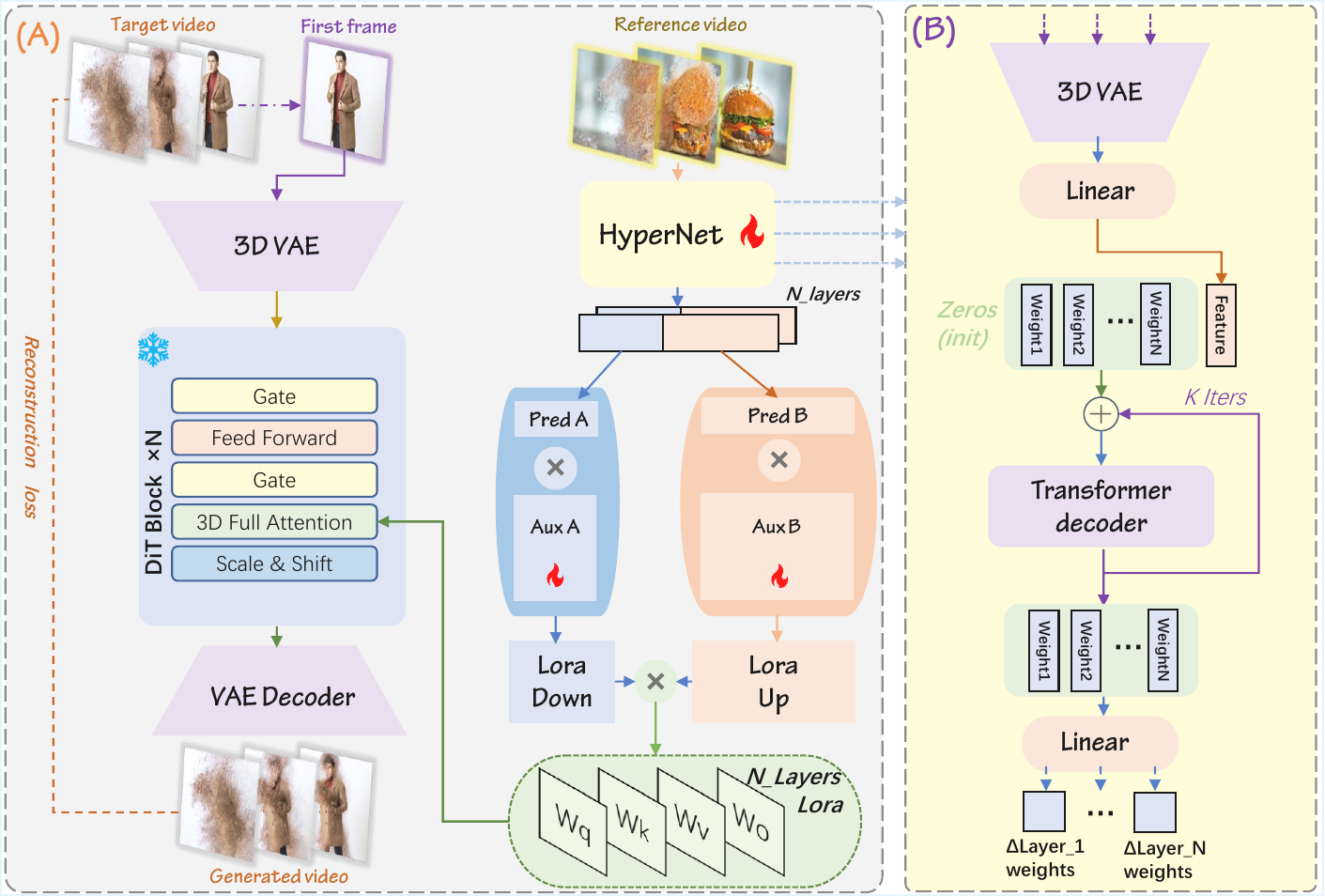}
\caption{
Overview of the proposed \textbf{Video2LoRA} framework. A reference semantic video is first fed into the HyperNetwork, where a 3D-VAE encoder extracts spatio-temporal latent features that are linearly projected into the layer-wise LightLoRA subspaces. The projected features are concatenated with zero-initialized weight tokens and processed by a Transformer decoder, which iteratively predicts the LightLoRA components $(A_{\text{pred}}, B_{\text{pred}})$ for each diffusion layer. These predicted components are then fused with the trainable auxiliary matrices $(A_{\text{aux}}, B_{\text{aux}})$ to form the final semantic-specific LoRA weights. The resulting LoRA adapters are injected into the frozen DiT backbone and optimized end-to-end with the vanilla diffusion loss, enabling semantic-controllable video generation from reference videos.
}
\label{fig:frame}
\end{figure*}

\section{Introduction}
\label{sec:intro}
The rapid evolution of generative AI has profoundly transformed visual content creation, enabling unprecedented levels of efficiency, controllability, and expressiveness. In particular, large-scale pretrained video diffusion models~\cite{zheng2024open,yang2024cogvideox,wan2025wan,kong2024hunyuanvideo} have exhibited impressive capabilities in semantic comprehension and temporally coherent synthesis. Recent advances in controllable video generation~\cite{wang2025cinemaster,wang2025multishotmaster,ma2024followpose,ma2025followyourmotion} have predominantly focused on spatially aligned paradigms,leveraging modalities such as depth maps~\cite{peng2024controlnext,wang2025cinemaster}, human poses~\cite{hu2024animate}, edge sketches~\cite{geng2025motion}, keypoints~\cite{gu2025diffusion,jeong2025track4gen}, or optical flow~\cite{jin2025flovd} to impose spatially consistent guidance during generation. Unified frameworks operating under these pixel-aligned conditions have been investigated~\cite{jiang2025vace}, demonstrating stable and precise control over structural attributes. In contrast, semantic-controlled video generation that encompasses aspects such as visual effects~\cite{liu2025vfx,mao2025omni,li2025vfxmaster,bai2025semanticgen}, camera motion~\cite{bai2025recammaster}, and personalized styles remains~\cite{ye2025stylemaster} relatively underexplored despite its close alignment with real-world creative demands. Such high-level controls are inherently more intuitive to human users but challenging to formalize or acquire, as they often lack explicit spatial or parametric representations (e.g., camera trajectories or semantic annotations). Consequently, establishing a unified, generalizable, and user-friendly framework for semantic video control remains an open and pressing challenge.

Existing controllable video generation methods exhibit limited scalability and generalization due to their condition-specific designs. A prevalent line of research fine-tunes either the diffusion backbone or dedicated Low-Rank Adapter (LoRA)~\cite{hu2022lora} for each semantic condition~\cite{liu2025vfx,mao2025omni}, effectively memorizing condition-specific representations. Although such approaches yield satisfactory control within individual domains, they are computationally expensive, storage-inefficient, and fail to generalize across heterogeneous or composite semantics. Another research direction introduces task-specific architectures or inference branches customized for distinct control types, encoding prior knowledge directly into the model design~\cite{bai2025recammaster,ye2025stylemaster,zhang2025flexiact}. However, these handcrafted solutions lack interoperability and are inherently constrained to the semantics on which they are trained. As a result, existing paradigms remain fragmented, requiring substantial reconfiguration for new conditions and exhibiting poor zero-shot generalization to unseen semantic domains.

Inspired by the strong cross-domain personalization capability of hypernetworks demonstrated in HyperDreamBooth~\cite{ruiz2024hyperdreambooth}, we hypothesize that similar meta-adaptive mechanisms can endow video generation models with dynamic semantic adaptability. Specifically, hypernetworks~\cite{liu2026shine} can generate semantic-dependent lightweight parameters to modulate a frozen diffusion backbone, enabling flexible semantic conditioning without the need for task-specific retraining. Building upon this insight, we seek to extend the generalization ability of hypernetworks from personalized image synthesis to the broader domain of semantic video generation.

In this work, we introduce Video2LoRA, a unified and generalizable framework for semantic-controlled video generation that conditions on a reference video to synthesize semantically aligned outputs across both in-domain and out-of-domain scenarios. Video2LoRA achieves strong semantic adaptability through a novel hypernetwork-based generation paradigm, in which the hypernetwork predicts a set of lightweight LoRA weights, each less than 50 KB per semantic condition, and merges them with auxiliary matrices to form adaptive LoRA modules injected into a frozen diffusion backbone. This design enables the model to dynamically modulate generation behavior according to diverse semantic cues while requiring no per-condition finetuning. Unlike HyperDreamBooth~\cite{ruiz2024hyperdreambooth}, which relies on pre-trained personalized weights as supervision and a three-stage training pipeline with rank relaxation, Video2LoRA is trained end-to-end in a single stage using only the diffusion loss. Our approach eliminates the need for any pre-training or fine-tuning phases, enabling the hypernetwork to directly learn and generalize semantic representations from raw video data without explicit supervision. Extensive experiments on both in-domain and out-of-domain benchmarks demonstrate that Video2LoRA achieves high-fidelity, semantically aligned video generation under diverse control conditions while exhibiting strong generalization to unseen semantics. Our contributions are summarized as follows:

\begin{itemize}
    \item \textbf{Lightweight LoRA representation.} We propose a compact LoRA formulation by training the video generation model within a low-dimensional, trainable weight subspace constructed from a random orthogonal incomplete basis in the low-rank adaptation space. Each semantic condition requires less than 50 KB of parameters.
    \item \textbf{Novel hypernetwork architecture.} We design a novel hypernetwork that leverages the lightweight LoRA configuration to dynamically predict semantic-specific LoRA components for a given video condition, enabling efficient and adaptive control within a unified diffusion backbone.
    \item \textbf{End-to-end semantic generalization.} Unlike prior approaches that rely on pre-trained semantic weights or explicit supervision for each condition, Video2LoRA trains the hypernetwork directly using diffusion objectives, allowing it to implicitly capture semantic relationships and generalize to unseen conditions.
\end{itemize}

\section{Related work}
\label{sec:Related work}

\subsection{Video Generation}
Recent advances in video generation have been largely driven by diffusion models~\cite{ho2020denoising,nichol2021improved, ma2025followcreation, ma2026fastvmt, ma2025followfaster, ma2024followyouremoji, wang2024cove, ma2025followyourclick, long2025follow, wang2024taming, feng2025dit4edit, chen2025contextflow}, particularly those adopting Diffusion Transformer (DiT) architectures~\cite{peebles2023scalable}, which integrate the generative strength of diffusion processes with the contextual modeling power of transformers~\cite{vaswani2017attention}. Such designs greatly enhance temporal coherence and improve motion dynamics. For example, Open-Sora~\cite{zheng2024open} demonstrates efficient long-duration synthesis through scalable transformer blocks and optimized spatiotemporal attention; CogVideoX~\cite{yang2024cogvideox} employs full 3D self-attention to jointly model spatial-temporal dependencies, significantly improving frame-to-frame consistency; and Wan 2.2~\cite{wan2025wan} incorporates a Mixture-of-Experts (MoE) design to achieve scalable specialization across heterogeneous video content. Despite these advances, most pre-trained DiTs~\cite{wang2025characterfactory,wang2025stableidentity,ma2022visual,ma2025controllable} remain limited to text-only or frame-based conditioning, restricting fine-grained semantic control. To address this limitation, recent efforts introduce task-specific modules or customized inference mechanisms to enable more flexible and user-driven video generation.

\subsection{Controllable Video Generation}
Controllable video generation can be broadly categorized into spatial-alignment and semantic-control paradigms:

The spatial-alignment paradigm leverages explicit structural cues, such as depth~\cite{peng2024controlnext,wang2025cinemaster}, pose~\cite{hu2024animate}, mask~\cite{bian2025videopainter}, optical flow~\cite{jin2025flovd}, or motion trajectories~\cite{geng2025motion} to impose pixel-level constraints on synthesis. LongVie~\cite{gao2025longvie} integrates multimodal depth and keypoint guidance to achieve temporally coherent ultra-long video generation. Animate Anyone~\cite{hu2024animate} employs pose-based conditioning with spatial attention to achieve appearance-consistent character animation. Motion Prompting~\cite{geng2025motion} introduces trajectory-based motion cues (“motion prompts”) to flexibly control object and camera dynamics. While these methods excel at structure-aware control and fine-grained synthesis, they depend on labor-intensive signal extraction or external annotations, making them less suitable for abstract or semantic-level conditioning.

The semantic-control paradigm involves high-level, concept-driven manipulations such as visual effects, camera motion (e.g., trajectories, zooms, or orbits), and personalized stylization (e.g., Ghibli, anime, or Minecraft styles). Existing approaches, such as VFXCreator~\cite{liu2025vfx} and GS-DiT~\cite{bian2025gs}, achieve control by fine-tuning diffusion backbones or condition-specific Low-Rank Adapters (LoRA) for each semantic condition, including motion type, visual style, or camera behavior, thereby “memorizing” domain-specific representations. Although effective within isolated domains, these methods incur substantial computational cost, suffer from poor parameter efficiency, and fail to generalize across heterogeneous or compositional semantics.
Other works, including StyleMaster~\cite{ye2025stylemaster}, DiTFlow~\cite{pondaven2025video}, and VD3D~\cite{bahmani2024vd3d}, introduce task-specific architectures for style extraction, motion guidance, or camera-based 3D reasoning, embedding prior knowledge directly into the model structure. However, such specialization inherently limits flexibility and hinders generalization to unseen semantics. OmniEffects~\cite{mao2025omni} attempts to integrate multiple video semantics via a Mixture-of-Experts (MoE) framework but remains confined to in-domain compositions without achieving true cross-domain adaptability. To overcome these limitations, we propose Video2LoRA, a unified semantic video generation framework that enables zero-shot generalization across diverse semantic conditions..

\section{Method}

We introduce Video2LoRA, a unified and generalizable framework for controllable video generation that learns semantic control end-to-end by dynamically producing lightweight LoRA parameters for any video semantic. Unlike prior approaches~\cite{liu2025vfx} that depend on pre-trained semantic experts or condition-specific finetuning pipelines, Video2LoRA directly adapts a diffusion-based video backbone using semantic cues extracted from reference videos, enabling stronger flexibility, scalability, and zero-shot generalization.

Our method is built upon three core components.
First, Sec.~\ref{sec:pre} reviews the fundamentals of the CogVideoX diffusion backbone and LoRA-based parameter-efficient adaptation.
Then, Sec.~\ref{sec:lilora} introduces our \emph{LightLoRA} representation—a compact and trainable low-dimensional parameterization that enables the HyperNetwork to generate semantic-adaptive LoRA weights efficiently.
Next, Sec.~\ref{sec:hyperNet} describes the proposed Transformer-based HyperNetwork that predicts these semantic-dependent LoRA components by analyzing spatio-temporal features extracted from a reference video.
Finally, Sec.~\ref{sec:Video2LoRA} presents the full end-to-end training pipeline, where predicted LoRA weights, auxiliary matrices, and the CogVideoX backbone are jointly optimized under the standard image-to-video diffusion objective.

\subsection{Preliminaries}
\label{sec:pre}
\textbf{Video Generation Backbone.} 
We build our framework upon \textbf{CogVideoX-5B-I2V}~\cite{yang2024cogvideox}, an image-to-video diffusion model that integrates a \textbf{3D Variational Autoencoder (VAE)}~\cite{kingma2013auto}, a \textbf{Diffusion Transformer (DiT)} backbone, and a \textbf{T5-based text encoder}~\cite{raffel2020exploring}. Given an input image $I \in \mathbb{R}^{h \times w \times c}$ and a textual prompt, the model synthesizes a video $V \in \mathbb{R}^{f \times h \times w \times c}$ containing $f$ frames. 

During training, the 3D VAE compresses each target video into latent representations that capture spatial-temporal structures, while the first frame is separately encoded for temporal alignment. These latent features are concatenated and passed through the DiT backbone, which iteratively refines the noisy latent sequence through a diffusion-based denoising process guided by text embeddings. This diffusion formulation enables CogVideoX to learn coherent temporal dynamics and high-fidelity visual content, serving as a strong foundation for our controllable video generation framework.

\textbf{Low-Rank Adaptation (LoRA)}~\cite{hu2022lora} provides a parameter-efficient fine-tuning strategy that has been widely adopted in diffusion-based generation frameworks. Instead of updating all network parameters, LoRA optimizes low-rank residual matrices that are added to the frozen model weights. Formally, for a given layer $l$ with a weight matrix $W \in \mathbb{R}^{n \times m}$, LoRA introduces a learnable residual term:
\begin{equation}
\Delta W = AB,
\end{equation}
where $A \in \mathbb{R}^{n \times r}$ and $B \in \mathbb{R}^{r \times m}$, with rank $r \ll \min(n, m)$. 
This low-rank decomposition significantly reduces the number of trainable parameters while maintaining expressive adaptation capacity.


\subsection{Light Weight Lora Representation}
\label{sec:lilora}
To enable direct generation of semantic-specific weight subsets through a HyperNetwork while maintaining semantic fidelity, editability, and generalization, we propose a novel low-dimensional trainable weight space for semantic control in video diffusion models. This compact formulation enables multi-semantic LoRA models that are over \textbf{150$\times$ smaller} than the CogVideoX backbone and more than \textbf{20$\times$ smaller} than single-semantic LoRA-CogVideoX variants.

The core idea of our \textit{LightLoRA} is to further decompose the rank-1 LoRA weight space while maintaining trainability of the decomposed factors. As illustrated in Figure~\ref{fig:frame}(A), this can be understood as decomposing the \texttt{Down(A)} and \texttt{Up(B)} matrices in Eq.~(2) into two components:
\begin{equation}
A = A_{\text{aux}} A_{\text{pred}}, \quad B = B_{\text{pred}} B_{\text{aux}},
\end{equation}
where $A_{\text{aux}} \in \mathbb{R}^{n\times a}$ and $B_{\text{aux}} \in \mathbb{R}^{b\times m}$ are auxiliary matrices initialized with row-wise orthogonal vectors of constant magnitude and set to be trainable. The matrices $A_{\text{pred}} \in \mathbb{R}^{a\times r}$ and $B_{\text{pred}} \in \mathbb{R}^{r\times b}$ are dynamically predicted by the HyperNetwork for each semantic condition. Consequently, the residual weight in each linear layer is expressed as:
\begin{equation}
\Delta W x = A_{\text{aux}} A_{\text{pred}} B_{\text{pred}} B_{\text{aux}},
\end{equation}
where $r \ll \min(n,m)$, $a < n$, and $b < m$. Two newly-introduced hyperparameters, $a$ and $b$, control the dimensionality of the auxiliary subspace. In our experiments, setting $a=100$ and $b=50$ yields only \textbf{23K trainable variables} (approximately \textbf{30 KB} in \texttt{bf16}), while preserving strong semantic adaptability and zero-shot generalization. Unlike LiDB~\cite{ruiz2024hyperdreambooth}, where the auxiliary matrices are frozen, our trainable $A_{\text{aux}}$ and $B_{\text{aux}}$ act as \textit{semantic priors} that encode generalizable video semantics. During training, the HyperNetwork learns to combine these priors via the dynamically predicted $A_{\text{pred}}$ and $B_{\text{pred}}$, producing condition-specific rank-1 LoRA adapters. 


\begin{figure*}[t] 
\centering 
\includegraphics[width=1\linewidth, height=0.65\linewidth, keepaspectratio=false]{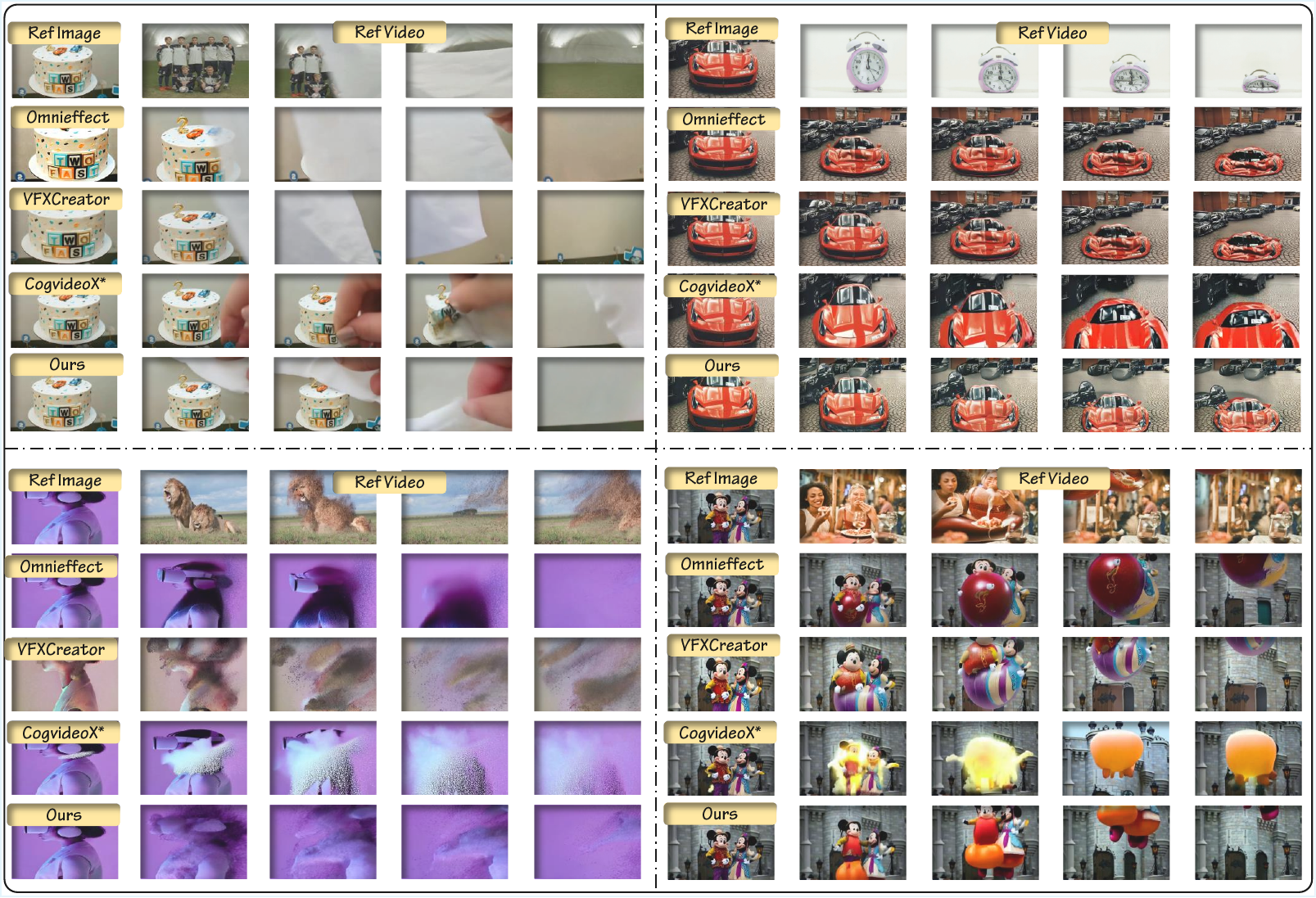}
\caption{Qualitative comparison with VFXCreator~\cite{liu2025vfx} and Ominieffect~\cite{mao2025omni} on the OpenVFX dataset. CogVideoX* refers to the CogVideoX model after supervised fine-tuning on our dataset.} 
\label{fig:experiments} 
\end{figure*}

\subsection{HyperNetwork Architecture}
\label{sec:hyperNet}
As illustrated in Figure~\ref{fig:frame}(B), the proposed HyperNetwork $\mathcal{H}_{\eta}$ is composed of a 3D-VAE encoder, a linear projection layer, and a Transformer-based decoder. The encoder shares the same architecture as the video backbone’s 3D-VAE to ensure feature-level alignment between the adaptation module and the generative model. Since the encoder is sequentially dependent on the weights of different layers, effective model personalization requires capturing inter-layer dependencies in the generated LoRA parameters. Previous works~\cite{liu2025vfx} overlook this dependency, treating layer weights as conditionally independent. In contrast, our Transformer decoder explicitly models these positional dependencies through learned positional embeddings, analogous to how language models capture contextual relationships among tokens. This design enables the HyperNetwork to reason over structured relationships between layers rather than generating them independently.

The encoder first extracts spatio-temporal latent features $f$ from the input reference video, capturing both motion dynamics and semantic content. These features are projected through a linear layer and passed into the Transformer decoder, which sequentially predicts the semantic-specific LoRA components $(A_{pred}, B_{pred})$ across layers.

To further enhance inter-layer consistency, we adopt an iterative refinement mechanism similar to recurrent inference. At each iteration $k$, the decoder refines its prediction based on the previous output:
\begin{equation}
\theta_{pred}^{(k)} = \mathcal{T}(f, \theta_{pred}^{(k-1)}),
\end{equation}
where $\mathcal{T}$ denotes the Transformer decoder and $\theta_{pred}^{(0)}$ is initialized to zero. The refinement process continues until $k=s$, where $s$ specifies the number of refinement steps. This iterative design effectively enforces semantic stability and temporal coherence while remaining computationally efficient, since the video encoding $f$ is computed only once and reused throughout the refinement process.

\subsection{HyperNetwork for Video Semantic Adaptation}
\label{sec:Video2LoRA}

To enable fast and unified semantic control in video generation, we adopt a HyperNetwork-based adaptation mechanism for image-to-video diffusion models. A HyperNetwork $\mathcal{H}_{\eta}$, parameterized by $\eta$, takes the 3D-VAE feature representation $x_i$ of a reference semantic video as input and predicts the low-rank \textit{LightLoRA} residuals $\hat{\theta} = \mathcal{H}_{\eta}(x_i)$. Each predicted $\hat{\theta}$ corresponds to the LoRA component of a specific attention layer and is fused with the auxiliary matrices defined in Eq.~(3) to construct the final adaptive LoRA weights. These LoRA adapters are injected into the frozen DiT backbone of CogVideoX-I2V, enabling semantic conditioning within a single diffusion training stage.

In contrast to prior personalization methods~\cite{ruiz2024hyperdreambooth,liu2025vfx,mao2025omni} that require pre-optimized semantic weights or condition-specific finetuning, our framework jointly trains both the HyperNetwork and the auxiliary matrices solely under the standard I2V diffusion objective. This unified learning scheme allows the HyperNetwork to absorb semantic priors directly from diffusion dynamics, yielding strong generalization across diverse in-domain and out-of-domain semantic conditions.

During training, the 3D-VAE encodes the target semantic video into latent features $z$, while the first frame is replaced by a placeholder token ($-1$) for temporal alignment and encoded as $z_i$. The concatenated latent pair $[z_i, z]$ is then passed into the DiT denoiser, where the HyperNetwork-predicted LoRA weights and auxiliary matrices jointly modulate the denoising process. The denoising network $\epsilon_{\Theta}$ is trained using the standard diffusion objective:
\begin{equation}
\mathcal{L}_{\text{diff}}(\Theta) = 
\mathbb{E}_{t, z, \epsilon}
\left[
\| \epsilon - \epsilon_{\Theta}(z_t, t, g) \|_2^2
\right],
\end{equation}
where $\epsilon \sim \mathcal{N}(0, I)$ is Gaussian noise, $z_t$ is the noisy latent at timestep $t$, $g$ is the text embedding, and $\Theta$ denotes the parameters of the denoising model. This diffusion-driven supervision propagates gradients through the injected LoRA modules, effectively training the HyperNetwork and auxiliary matrices in an end-to-end manner. Figure~\ref{fig:frame} provides an overview of the full training pipeline.

\section{Experiments}
\begin{table*}[!t]
\caption{\textbf{Performance comparison on OpenVFX dataset.} CogvideoX* refers to CogVideoX after supervised fine-tuning on our dataset. Avg. represents the average score over all effects. The highest metric values are highlighted in \textbf{bold}. }
\centering
\footnotesize
\setlength\tabcolsep{1pt}
\begin{tabularx}{\textwidth}{@{}cccccccccccccccccc@{}}
\toprule
\textbf{Metrics} & \textbf{Methods} & \textbf{Cake} & \textbf{Crumble} & \textbf{Crush} & \textbf{Decap} & \textbf{Deflate} & \textbf{Dissolve} & \textbf{Explode} & \textbf{Eye-pop} & \textbf{Harley} & \textbf{Inflate} & \textbf{Levitate} & \textbf{Melt} & \textbf{Squish} & \textbf{Ta-da} & \textbf{Venom} & \textbf{Avg.} \\ 
\midrule

\multirow{4}{*}{\textbf{FVD$\downarrow$}} 
& CogvideoX* & 1732 & 1849 & 1195 & 1937 & 1664 & 1916 & 2427 & 1649 & \textbf{2232} & 2169 & 1473 & 2941 & 1938 & 1431 & \textbf{2792} & 1956 \\
& VFX Creator & 1776 & 1580 & 1156 & 1754 & 1997 & 1607 & 1886 & \textbf{1447} & 2815 & 2089 & 1143 & 2547 & 1880 & 1107 & 3062 & 1856 \\
& Omini-Effects & \textbf{1548} & 1410 & 1136 & \textbf{1263} & \textbf{1037} & 1543 & 2044 & 1559 & 2501 & 1464 & 1295 & 2418 & 1923 & 1368 & 2678 & 1679 \\
 \cmidrule(){2-18}
& Ours & 1573 & \textbf{1358} & \textbf{1107} & 1677 & 1294 & \textbf{1412} & \textbf{1125} & 1528 & 2466 & \textbf{1162} & \textbf{1005} & \textbf{2193} & \textbf{1606} & \textbf{1027} & 2973 & \textbf{1568} \\ 
\midrule

\multirow{4}{*}{\textbf{\makecell[c]{Dynamic\\Degree}$\uparrow$}}
& CogvideoX* & 1.0 & 1.0 & 0.6 & 0.6 & 0.4 & 0.4 & 1.0 & 0.0 & 1.0 & 0.4 & 0.0 & 0.6 & 1.0 & 0.8 & 1.0 & 0.65 \\
& VFX Creator & 1.0 & 1.0 & 0.0 & 0.6 & 0.0 & \textbf{0.8} & 1.0 & 0.0 & 1.0 & 1.0 & 0.0 & 0.6 & 1.0 & 1.0 & 1.0 & 0.67 \\
& Omini-Effects & 1.0 & 1.0 & 0.6 & 0.6 & 0.2 & 0.4 & 1.0 & 0.2 & 1.0 & 1.0 & 0.0 & \textbf{0.8} & 1.0 & 0.8 & 1.0 & 0.71 \\ 
 \cmidrule(){2-18} 
& Ours & \textbf{1.0} & \textbf{1.0} & \textbf{0.8} & 0.6 & \textbf{0.6} & \textbf{0.8} & \textbf{1.0} & \textbf{0.2} & \textbf{1.0} & \textbf{1.0} & \textbf{0.2} & 0.6 & \textbf{1.0} & \textbf{1.0} & \textbf{1.0} & \textbf{0.78} \\ 
\midrule

\multirow{4}{*}{\textbf{\makecell[c]{Motion\\Smoothness}$\uparrow$}}
& CogvideoX* & 97.25 & 96.80 & 98.10 & 97.95 & 98.42 & 98.15 & 97.83 & 98.67 & \textbf{99.02} & 98.45 & 98.76 & 98.01 & 97.56 & 98.05 & 97.48 & 98.17 \\
& VFX Creator & 97.84 & 97.10 & 98.23 & 97.68 & 98.51 & \textbf{98.60} & 97.96 & 98.72 & 98.88 & 98.32 & 98.69 & 98.14 & 97.70 & 98.26 & 97.62 & 98.16 \\
& Omni-Effects & 97.66 & \textbf{97.58} & 98.34 & 97.83 & 98.47 & 98.41 & \textbf{98.22} & 98.69 & 98.95 & 98.48 & 98.71 & 98.25 & 97.92 & 98.30 & 97.73 & 98.24 \\
\cmidrule(){2-18}
& Ours & \textbf{98.02} & 97.34 & \textbf{98.56} & \textbf{99.24} & \textbf{99.28} & 98.42 & 98.06 & \textbf{99.39} & 97.04 & \textbf{99.01} & \textbf{99.55} & \textbf{98.46} & \textbf{98.30} & \textbf{98.48} & \textbf{96.51} & \textbf{98.50} \\
\midrule

\multirow{4}{*}{\textbf{\makecell[c]{Aesthetic\\Quality}$\uparrow$}}
& CogvideoX* & 0.49 & 0.52 & 0.50 & 0.47 & 0.55 & 0.52 & 0.46 & 0.54 & 0.50 & 0.47 & 0.54 & 0.51 & 0.49 & 0.53 & 0.48 &0.506 \\
& VFX Creator & 0.50 & 0.51 & 0.53 & 0.46 & 0.57 & 0.54 & 0.49 & \textbf{0.57} & 0.56 & 0.48 & 0.52 & 0.50 & 0.47 & 0.55 & 0.50 & 0.519 \\
& Omni-Effects & 0.52 & 0.54 & 0.55 & 0.49 & 0.58 & 0.56 & 0.51 & 0.54 & \textbf{0.58} & \textbf{0.53} & 0.55 & \textbf{0.52} & 0.48 & 0.57 & 0.52 &0.537 \\
\cmidrule(){2-18}
& Ours & \textbf{0.58} & \textbf{0.59} & \textbf{0.57} & \textbf{0.55} & \textbf{0.59} & \textbf{0.58} & \textbf{0.54} & 0.56 & 0.57 & 0.52 & \textbf{0.58} & 0.51 & \textbf{0.53} & \textbf{0.59} & \textbf{0.55} & \textbf{0.565} \\

\bottomrule
\end{tabularx}
\label{tab:in-domain}
\vspace{-0.3cm}
\end{table*}

\subsection{Implementation Details}

We employ \textbf{CogVideoX-I2V-5B}~\cite{yang2024cogvideox} as the frozen backbone for all experiments. During training, the \textit{LightLoRA} weights predicted by the HyperNetwork are combined with the auxiliary matrices to form rank-1 low-rank adapters, which are injected into the 3D Transformer blocks of the backbone. The Video2LoRA framework is trained on the \textbf{4k} dataset and randomly pairs samples from the same semantic category to construct reference-target video pairs. Each video is uniformly sampled to 49 frames at 8~fps and resized to a resolution of $480\times720$ pixels. The reference videos are zero-padded to match the spatial and temporal dimensions of the target videos. We use the \textbf{AdamW}~\cite{kingma2014adam} optimizer with a learning rate of $1\times10^{-4}$ and train only the parameters of the HyperNetwork and auxiliary matrices while keeping the backbone frozen. Training is performed on 8 NVIDIA A800 GPUs for approximately 20K iterations.

\subsection{Datasets}

Our training dataset is constructed from multiple sources, including the open-source \textbf{Open-VFX}~\cite{liu2025vfx} dataset, commercial video platforms such as \textbf{Higgsfield}~\cite{higgsfield2025} and \textbf{PixVerse}~\cite{pixverse2025}, as well as publicly available online resources. In total, the dataset comprises approximately \textbf{4K video samples} spanning over \textbf{200 distinct semantic categories}, covering a diverse range of effects, including character transformations, environmental transitions, camera motion dynamics, object stylization, and artistic style variations. 

To further evaluate the robustness and generalization capability of our framework, we curate a dedicated \textbf{out-of-domain (OOD)} test set containing unseen semantic conditions. This dataset enables a systematic assessment of the model’s ability to adapt to novel visual effects and semantic distributions beyond the training domain.

\begin{figure}
    \centering
    \includegraphics[width=1.0\linewidth]{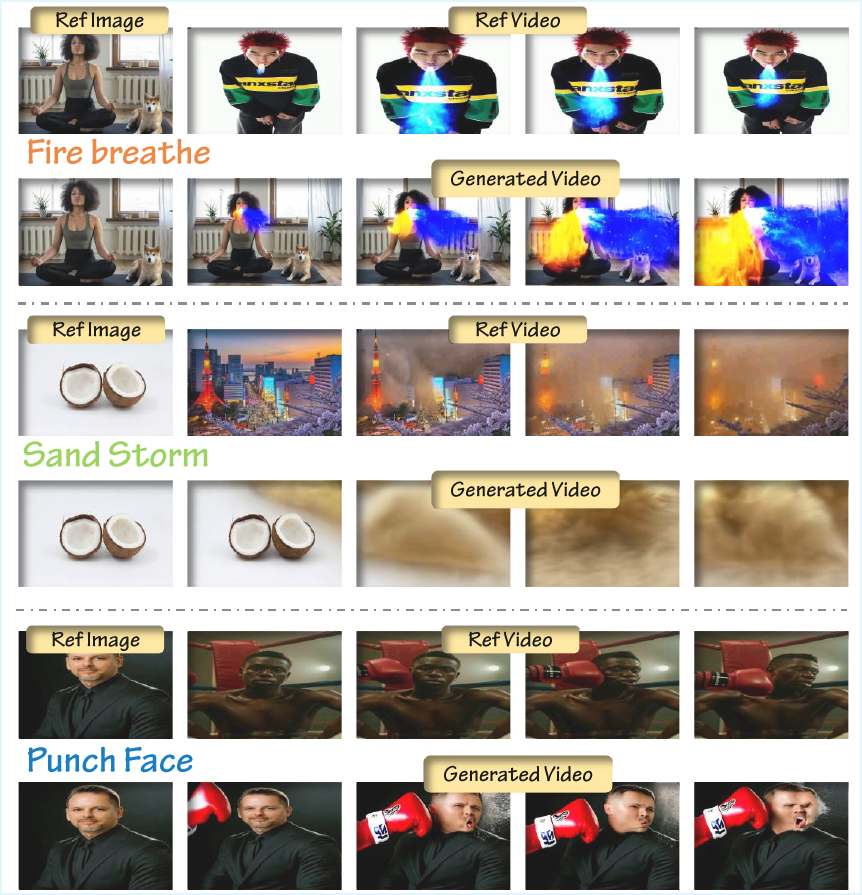}
    \caption{Out-of-Domain Comparison}
    \label{fig:OOD}
\end{figure}
\subsection{Evaluation Metrics}
\begin{table*}[t]
\centering
\caption{\textbf{Ablation study and zero shot generationon Video2LoRA.} We evaluate the impact of different iterative steps $k$ and auxiliary matrix settings $(a,b)$ on performance. The upper part compares different $k$ values, while the lower part analyzes the influence of $(a,b)$.}
\resizebox{0.7\textwidth}{!}{
\begin{tabular}{lcccc}
\toprule
\textbf{Methods} & \textbf{FVD$\downarrow$} & \textbf{Dynamic Degree$\uparrow$} & \textbf{Motion Smoothness$\uparrow$} & \textbf{Aesthetic Quality$\uparrow$} \\
\midrule
Ours & \textbf{1358} & \textbf{0.72} & \textbf{98.50} & \textbf{0.57} \\
Ours (Zero-Shot) & 1492 & 0.71 & 98.37 & 0.54 \\
\midrule
Ours ($k$=1) & 1764 & 0.63 & 97.45 & 0.51 \\
Ours ($k$=2) & 1598 & 0.67 & 97.92 & 0.53 \\
Ours ($k$=8) & 1439 & 0.70 & 98.23 & 0.55 \\
\midrule
Ours ($a$=60, $b$=30) & 1512 & 0.69 & 98.10 & 0.54 \\
Ours ($a$=160, $b$=80) & 1384 & 0.71 & 98.42 & 0.56 \\
\bottomrule
\end{tabular}}
\label{tab:ablation}
\vspace{-0.3cm}
\end{table*}

Following previous studies~\cite{liu2025vfx,mao2025omni,bian2025videopainter}, we comprehensively evaluate the proposed method using multiple quantitative metrics, including FVD~\cite{unterthiner2018towards}, dynamic degree~\cite{teed2020raft}, motion smoothness~\cite{raffel2020exploring}, and aesthetic quality~\cite{schuhmann2022laion}. These metrics collectively reflect different aspects of video generation performance, and detailed descriptions are omitted for brevity.

To quantitatively evaluate the in-domain performance of our approach, we conduct experiments on 15 semantic categories selected from the \textbf{Open-VFX} test set. As shown in Table~\ref{tab:in-domain}, we comprehensively compare \textbf{Video2LoRA} against two state-of-the-art VFX generation methods and a baseline model fine-tuned on the same dataset. The results demonstrate that our \textbf{Video2LoRA} consistently outperforms all competing approaches in terms of average scores across all evaluation metrics, exhibiting superior visual fidelity, motion coherence, aesthetic appeal, and dynamic range. Notably, for complex effects involving particle dynamics or strong subject interactions, such as \textit{Crumble}, \textit{Crush}, \textit{Decap}, and \textit{Inflate}, our model achieves substantially higher realism and temporal consistency. These findings indicate that \textbf{Video2LoRA} not only learns to capture semantic information from reference videos but also reproduces such semantics with higher fidelity and stability over time.

To further evaluate the model's generalization capability beyond the training domain, we conduct a zero-shot out-of-domain (OOD) evaluation, with results reported in Table~\ref{tab:ablation}. The evaluation shows that the model's performance on unseen videos is comparable to that observed in the in-domain setting, demonstrating that \textbf{Video2LoRA} can generate high-quality, temporally coherent videos even for previously unseen semantic effects. These results further validate the framework's robust zero-shot generalization and semantic adaptation capabilities.

\subsection{Qualitative Comparison}
We conduct a qualitative comparison between \textbf{Video2LoRA} and three representative models across four distinct visual effect categories, as illustrated in Figure~\ref{fig:experiments}. Compared with the fine-tuned \textbf{CogVideoX-5B} and two state-of-the-art VFX generation frameworks, \textbf{Video2LoRA} produces results with notably higher visual fidelity and semantic accuracy. For instance, under the \textit{Dissolve} effect, our model not only captures the gradual disintegration of the subject with fine-grained temporal consistency but also realistically simulates secondary physical behaviors, such as the natural fall of the subject’s VR headset after dissolution. Similarly, for the \textit{Levitate} effect, \textbf{Video2LoRA} generates smooth and coherent motion trajectories while maintaining semantic alignment with the reference. 

Furthermore, Figure~\ref{fig:OOD} presents the results of zero-shot experiments. Even on unseen videos, \textbf{Video2LoRA} generates content whose visual style and semantic effects are well-aligned with the reference videos, accurately capturing the intended semantics. For example, in the \textit{Punch Face} effect, the model successfully generates the entire reaction process of a punch to the face, including precise facial deformations and realistic motion of fluids, demonstrating high-fidelity motion and semantic accuracy. These results highlight the model's strong zero-shot generalization capability. Overall, \textbf{Video2LoRA} achieves visual quality and semantic coherence that match or surpass current open-source state-of-the-art models, emphasizing its superior ability in precise controllable video effect synthesis.

\subsection{Ablation Study}

We first analyze the influence of the \textit{LightLoRA} configuration by varying the dimensions of the predicted matrices $(A_{\text{pred}}, B_{\text{pred}})$, which are controlled by hyperparameters $a$ and $b$. Specifically, we experiment with three settings: $( 60,30)$, $(100,50)$, and $(160,80)$. As shown in Table~\ref{tab:ablation}, the configuration $(100,50)$ achieves the best overall video generation quality. The smallest setup $(60,30)$ fails to capture sufficient semantic diversity due to its limited representational capacity, while increasing the parameter size to $(160,80)$—a 1.6× increase—does not lead to further improvements and even slightly degrades performance, likely due to overfitting and reduced semantic sparsity. These results indicate that an appropriately compact latent LoRA space is crucial for effective semantic adaptation.

We further investigate the effect of the iterative prediction mechanism introduced in Sec.~\ref{sec:Video2LoRA}. Specifically, we compare our full model (with $k=4$ refinement steps) against variants with fewer iterations ($k=1$, $k=2$) and a deeper refinement setup ($k=8$). As summarized in Table~\ref{tab:ablation}, the performance improves as the number of iterations increases up to $k=4$, beyond which further refinement yields diminishing returns and slightly higher computational cost. This demonstrates that four refinement rounds provide an optimal balance between semantic consistency, stability, and efficiency.

\section{Conclusion}

In this work, we introduce \textbf{Video2LoRA}, a unified  framework for semantic-controlled video generation that leverages a hypernetwork to predict semantic-specific LoRA weights from a reference video. By decoupling semantic adaptation from backbone modification, freezing the diffusion model while only training a compact hypernetwork and auxiliary matrices,
our approach eliminates the need for per-condition fine-tuning or pre-trained adapters. This design enables strong zero-shot generalization to unseen semantic domains while maintaining high visual fidelity and temporal coherence. Extensive experiments on the Open-VFX dataset demonstrate that Video2LoRA outperforms existing methods across multiple metrics, including FVD, motion smoothness, and aesthetic quality, despite using significantly fewer parameters. Ablation studies further validate the effectiveness of our lightweight LoRA representation and iterative refinement strategy. We believe this paradigm opens a scalable path toward truly general-purpose semantic control in generative video models.

\section*{Acknowledgment}
This work was supported in part by the National Natural Science Foundation of China (Grant No. 82102135, 62472065, U23B2010), the Liaoning Province Science and Technology Joint Program (Grant No. 2024-MSLH-065), the Fundamental Research Funds for Central Universities (Grant No. DUT25Z2514, DUT24YG201).

{
    \small
    \bibliographystyle{ieeenat_fullname}
    \bibliography{main}
}


\end{document}